\RequirePackage[2020-02-02]{latexrelease}
\documentclass[twocolumn, switch]{article} % Method A for two-column formatting

\usepackage{preprint}
\usepackage{comment}
\usepackage{adjustbox}
\usepackage{multirow}

\usepackage{amsmath, amsthm, amssymb, amsfonts}
\usepackage{caption}
\usepackage{subcaption}

\usepackage[numbers,square]{natbib}
\usepackage[utf8]{inputenc}	% allow utf-8 input
\usepackage[T1]{fontenc}	% use 8-bit T1 fonts
\usepackage{xcolor}		% colors for hyperlinks
\usepackage[colorlinks = true,
            linkcolor = purple,
            urlcolor  = blue,
            citecolor = cyan,
            anchorcolor = black]{hyperref}	% Color links to references, figures, etc.
\usepackage{booktabs} 		% professional-quality tables
\usepackage{nicefrac}		% compact symbols for 1/2, etc.
\usepackage{microtype}		% microtypography
\usepackage{lineno}		    % Line numbers
\usepackage{float}			% Allows for figures within multicol
\usepackage{acronym}
\usepackage{lipsum}		%  Filler text

\usepackage{newfloat}
\DeclareFloatingEnvironment[name={Supplementary Figure}]{suppfigure}
\usepackage{sidecap}
\sidecaptionvpos{figure}{c}
\usepackage{graphicx}

\usepackage{titlesec}
\titlespacing\section{0pt}{12pt plus 3pt minus 3pt}{1pt plus 1pt minus 1pt}
\titlespacing\subsection{0pt}{10pt plus 3pt minus 3pt}{1pt plus 1pt minus 1pt}
\titlespacing\subsubsection{0pt}{8pt plus 3pt minus 3pt}{1pt plus 1pt minus 1pt}

\newacro{CNN}{Convolutional Neural Network}
\acrodefplural{CNN}{Convolutional Neural Networks}
\newacro{MLP}{Multilayer Perceptron}
\acrodefplural{MLP}{Multilayer Perceptrons}
\newacro{DNC}{Dynamic Node Correlation}
\newacro{AFA}{Adaptive Feature Aggregation}
\newacro{KNN}{K-nearest neighbor}
\newacro{ECC}{Edge-Conditioned Convolution}
\newacro{KNNG}{K-nearest neighbor graph}
\newacro{DNN}{Deep Neural Network}
\newacro{GNN}{Graph Neural Network}
\newacro{ReLU}{Rectified Linear Unit}
\newacro{AEC}{Architecture, Engineering, and Construction}
\newacro{ICP}{Iterative Closest Point}

\usepackage{tikz,xcolor,hyperref}

\definecolor{lime}{HTML}{A6CE39}
\DeclareRobustCommand{\orcidicon}{
	\begin{tikzpicture}
	\draw[lime, fill=lime] (0,0) 
	circle [radius=0.16] 
	node[white] {{\fontfamily{qag}\selectfont \tiny ID}};
	\draw[white, fill=white] (-0.0625,0.095) 
	circle [radius=0.007];
	\end{tikzpicture}
	\hspace{-2mm}
}
\foreach \x in {A, ..., Z}{\expandafter\xdef\csname orcid\x\endcsname{\noexpand\href{https://orcid.org/\csname orcidauthor\x\endcsname}
			{\noexpand\orcidicon}}
}
\title{Shrinking unit: a Graph Convolution-Based Unit for CNN-like 3D Point Cloud Feature Extractors}

\usepackage{xwatermark}
\newwatermark[firstpage,color=gray!90,angle=0,scale=0.28, xpos=0in,ypos=-5in]{*\, at2n19@soton.ac.uk - \dag\, bastian.plass@hs-mainz.de \\  \ddag\, thomas.klauer@hs-mainz.de}

\usepackage{authblk}

\author[1]{Alberto Tamajo * \orcidA{}}
\author[2]{Bastian Plaß \dag \orcidB{}}
\author[2]{Thomas Klauer \ddag \orcidC{}}
\affil[1]{Department of Electronics and Computer Science, University of Southampton}
\affil[2]{i3mainz, Institute for Spatial Information and Surveying Technology of Mainz University of Applied Sciences}

\begin{document}
\renewcommand{\sectionautorefname}{Section}
\renewcommand{\subsectionautorefname}{Subsection}
\renewcommand{\tableautorefname}{Table}
\renewcommand{\figureautorefname}{Figure}
\setlength\abovedisplayskip{4pt}
\setlength\belowdisplayskip{4pt}
\captionsetup[table]{skip=4pt}
\twocolumn[ % Method A for two-column formatting
  \begin{@twocolumnfalse} % Method A for two-column formatting
\maketitle

\begin{abstract}
3D point clouds have attracted increasing attention in architecture, engineering, and construction due to their high-quality object representation and efficient acquisition methods. Consequently, many point cloud feature detection methods have been proposed in the literature to automate some workflows, such as their classification or semantic segmentation. Nevertheless, the performance of point cloud automated systems significantly lags behind their image counterparts. While part of this failure stems from the irregularity, unstructuredness, and disorder of point clouds, which makes the task of point cloud feature detection significantly more challenging than the image one, we argue that a lack of inspiration from the image domain might be the primary cause of such a gap. Indeed, given the overwhelming success of \acp{CNN} in image feature detection, it seems reasonable to design their point cloud counterparts, but none of the proposed approaches closely resembles them. Specifically, even though many approaches generalise the convolution operation in point clouds, they fail to emulate the \acp{CNN} multiple-feature detection and pooling operations. For this reason, we propose a graph convolution-based unit, dubbed Shrinking unit, that can be stacked vertically and horizontally for the design of \ac{CNN}-like 3D point cloud feature extractors. Given that self, local and global correlations between points in a point cloud convey crucial spatial geometric information, we also leverage them during the feature extraction process. We evaluate our proposal by designing a feature extractor model for the ModelNet-10 benchmark dataset and achieve 90.64\% classification accuracy, demonstrating that our innovative idea is effective. Our code is available at \href{https://github.com/albertotamajo/Shrinking-unit}{github.com/albertotamajo/Shrinking-unit}.
\end{abstract}
\keywords{3D point clouds \and point cloud representation learning \and point cloud classification \and geometric deep learning \and convolutional neural networks} % (optional)
\vspace{0.35cm}

  \end{@twocolumnfalse} % Method A for two-column formatting
% Method A for two-column formatting
]
%\begin{multicols}{2} % Method B for two-column formatting (doesn't play well with line numbers), comment out if using method A

%%%%%%%%%%%%%%%  Main text   %%%%%%%%%%%%%%%
% \linenumbers
\section{Introduction}
A novel trend for the acquisition and reconstruction of real objects has been on the rise in the \ac{AEC} industry for several years. Indeed, while objects used to be manually captured and signalised by single representative points, surface scanning techniques are now leveraged to capture their shape, texture and volume information. This paradigm shift stems from the increasing accessibility and affordability of 3D surface scanning sensors, such as 3D scanners, (built-in) LiDAR sensors, and depth cameras \cite{Liang.2021, Plass.2021}, which make this technology available to a broader audience. For several reasons, 3D point clouds are the most popular data formats for 3D representations.
First, 3D point clouds represent the original geometric information in 3D Euclidean space without discretisation and information loss. Second, they allow bilateral modality transformation to 3D meshes and volumetric grids. Third, several point cloud operators, such as the \ac{ICP} algorithm \cite{Chen.1991, Besl.1992}, enable us to align different point cloud scans of complex acquisition geometries. However, despite being so rich in geometric information, high-level point cloud processing, such as classification and semantic segmentation, is far from straightforward. Indeed, standard deep neural networks cannot be easily adapted to the point cloud domain as they require data with a regular structure.
In contrast, point clouds are essentially irregular because they are a set of continuously distributed 3D points, and any permutation of the points' ordering does not change the spatial distribution. One of the earliest approaches to address these challenges relies on converting point cloud data into a set of 2D images and then processing them through \acp{CNN}. This approach, however, fails to provide accurate feature descriptors because the projection procedure usually discards substantial geometric information. A similar approach uses \acp{CNN} to extract features from point clouds' volumetric representation. The downside of this methodology is excessive memory usage and quantisation artifacts.
PointNet \cite{Qi.2017b} pioneers the direct application of deep learning on the raw point cloud by operating on each point independently and then applying a symmetric function to aggregate features. As a direct consequence, the relationships between neighbouring points are not explored during the feature detection process. Local structures, however, carry significant spatial information. To leverage these dependencies, some approaches propose to represent point clouds as graphs. In doing so, the edges in the graph explicitly represent the correlation between points which are then explored using graph convolution operations. Additionally, pooling operations
are usually applied to aggregate global features or incrementally reduce in size the point cloud inputs.  In spite of this, these graph-based methods are still far from emulating \acp{CNN}\cite{Lecun.1999, Lecun.2015}. This becomes evident when considering that the adopted pooling operations do not summarise the point clouds' local structures in the same way as the \acp{CNN}' pooling layers aggregate information from the images' local patches. Most strikingly, none of the proposed approaches detects multiple features at a given depth of the network as \acp{CNN} generate multiple feature maps. In this paper, we argue that, given the significant success shown by \acp{CNN} in image detection tasks, building \ac{CNN}-like 3D point cloud feature extractors should be given an attempt. To this end, here we make the following contributions:
\begin{enumerate}
    \item We propose a novel graph convolution-based unit, dubbed Shrinking unit, that leverages self and local correlation between points to extract features from point clouds. This unit is also equipped with a locality-based pooling procedure that shrinks the point cloud inputs by encoding local spatial geometric information. The latter operation emulates \acp{CNN} pooling operations.
    \item We propose to stack multiple Shrinking units vertically and horizontally for the development of \ac{CNN}-like 3D point cloud extractors. A horizontal stack enables the increasing reduction in size of the point clouds by progressively encoding their local structures, while vertical stacks detect multiple features at each layer of the network. Additionally, a horizontal stack allows us to capture long-range dependencies between points, which are also crucial for shape understanding. Experiments suggest this proposal to be effective in practice.
\end{enumerate}

The remainder of this paper is organised as follows. A description of state-of-the-art 3D point cloud feature extractors is given in \autoref{sec:state-of-the-art}. The details of the proposed Shrinking unit are described in \autoref{sec:shrinking_unit}. Experimental results and a comparison with state-of-the-art methods are provided in \autoref{sec:experiments}. Finally, \autoref{sec:conclusion} concludes this paper and offers an outlook on improvements and future research ideas.

\section{State of the art in 3D Point Cloud feature extraction}\label{sec:state-of-the-art}
Point clouds as an input modality present a unique set of challenges when building feature extractors. Indeed, their sparse and irregular structure makes it difficult to directly deploy the state-of-the-art approaches developed for 2D images. In order to overcome this difficulty, an abundance of methods has been crafted for the specific purpose of point cloud representation learning in a relatively short amount of time.
\subsection{Multiview-based methods}
\acp{CNN} \cite{Lecun.1999, Lecun.2015} can only process regular, ordered and structured data; thus, they are unsuitable for point cloud representation learning. To overcome this challenge, several approaches \cite{Su.2015, Leng.2015, Bai.2016, Kalogerakis.2017, Cao.2017, LeZhang.2018, yu2018multi, yang2019learning, feng2018gvcnn, wang2019dominant, ma2018learning, wei2020view}
convert point cloud data into a collection of 2D images and then process them through \acp{CNN}. Although some of these architectures have achieved good performance, even projecting a point cloud from multiple angles inevitably results in a loss of spatial geometric information. Consequently, multiview-based methods might overlook some salient features and thus are unreliable for the task of point cloud representation learning.
\subsection{Voxel-based methods}
Similarly to multiview-based methods, voxel-based methods \cite{Maturana.2015, Maturana.2015b, Qi.2016, Wang.2019c, Ghadai.2018, wu20153d, riegler2017octnet, le2018pointgrid, ben20173d} deploy \acp{CNN} to extract features from point cloud data. However, these approaches convert point clouds into fixed-size voxels instead of using multi-angle projections. In this way, the geometric information contained in point clouds is preserved as much as possible. Nevertheless, due to the sparse structure of point clouds, the voxelisation process can introduce artifacts. Specifically, a large number of invalid grids are generated in those point cloud locations that are very sparse. Furthermore, voxel-based methods suffer from high memory consumption. Hence, they are not suited for point cloud representation learning at a dense and accurate scale.
\subsection{Point-wise MLP methods}
Methods in this category model each point independently with shared \acp{MLP} and then use a symmetric aggregation function to aggregate global features.
PointNet \cite{Qi.2017b} is the pioneering work that proposed the direct application of deep learning on the raw point clouds. In more detail, PointNet's ability to operate directly on raw point cloud data relies on the use of two symmetric functions: (1) a shared \ac{MLP} and (2) a max-pooling function. The \ac{MLP} extracts a feature from each individual point, and the max-pooling function aggregates all the features into a global feature descriptor. However, even though this approach provides an efficient way for unstructured point cloud understanding, PointNet fails to capture local structures. Indeed, it processes each point independently without learning the relationships between neighbouring points. This is a significant shortcoming as points in a point cloud do not exist in isolation; rather, the aggregation of points and their local dependencies convey crucial spatial geometric information. In light of these considerations, several other methods, such as PointNet++ \cite{Qi.2017} and \cite{joseph2019momen, Zhao.2019, duan2019structural, lin2019justlookup}, improve in this regard, achieving better performance. Nevertheless, these approaches are essentially pointwise methods and hence still neglect correlations between neighbouring points.
\subsection{Graph-based methods}
Representing a point cloud as a graph
shows advantages over operating directly on the raw point
cloud data. Indeed, graphs can naturally deal with the sparse
and irregular structure of point clouds as well as explicitly
represent the correlation between points. To this end, graph-based approaches consider each point in a point cloud as a vertex of a graph and typically use variants of the \ac{KNN} method to generate the graph's directed edges. This structure is then leveraged for feature learning, typically through graph convolution and pooling operations. While the graph convolution operator captures local geometric information by convolving each point with respect to its neighbourhood, the pooling operation is adopted to either generate a global feature descriptor or a coarse graph. Hence, sequential graph pooling allows building deep architectures that produce multiple coarse graphs incrementally. This reduces in size the point cloud input, though preserving its salient properties, in the same spirit as \acp{CNN}.

As a pioneering work, \cite{simonovsky2017dynamic} constructs a graph by generating directed edges from each point to points lying within a fixed radius $p$. It adopts a multilayer architecture that alternates graph convolutions and graph pooling. The \ac{ECC} operator is proposed to convolve the graph's vertices, and graph pooling is implemented based on \cite{rusu20113d}. DGCNN \cite{Wang.2019} proposes a graph convolution operation, dubbed EdgeConv, to capture local neighbourhood information. The point cloud's graph is constructed using \ac{KNN}, and it is updated dynamically after each layer of the network. This multilayer system enables capturing higher-level representations incrementally and thus learning global correlations as well. Unlike \cite{rusu20113d}, DGCNN solely uses a global max-pooling operation at the end of the network instead of implementing sequential graph pooling. LDGCNN \cite{Zhang.2019} further improves DGCNN's performance by removing the transformation network and linking the hierarchical features from different layers. KCNet \cite{shen2018mining} forms a \ac{KNNG} to leverage neighborhood information for kernel correlation. After correlation learning, max-pooling is deployed to generate global feature descriptors of the point cloud inputs. G3D \cite{dominguez2018general} proposes a convolution filter as a variant of a polynomial of the adjacency matrix to explore local correlations in the point cloud's \ac{KNNG}. Pooling consists of multiplying the Laplacian and vertex matrix by a coarsening matrix. ClusterNet \cite{chen2019clusternet} extracts rotation-invariant features for each point from its $k$ nearest neighbours. An unsupervised agglomerative hierarchical clustering method with ward-linkage criteria \cite{mullner2011modern} constructs hierarchical structures, whose features are first learned through an EdgeConv operator and then aggregated using max-pooling.
AdaptConv \cite{Zhou.2021} proposes a novel convolution operation with an adaptive kernel. This design gives more convolution choices, making AdaptConv a very flexible solution. Furthest point sampling \cite{Qi.2017b} is used for graph pooling. 3D-GCN \cite{Lin.2020} pioneers a shift and scale invariant learnable kernel to capture local structural information. The point cloud's graph is reduced in size using random sampling. Point2Node \cite{Han.2021} employs a \ac{DNC} module that learns self, local and global correlations in a \ac{KNNG} sequentially and updates the nodes after each correlation learning to enhance their characteristics. Additionally, it makes use of an \ac{AFA} gate embedding to adaptively allows some channels of high-dimensional nodes to pass to the next correlation learning level while preventing others from proceeding.

Since \acp{CNN} achieve outstanding performance in image feature extraction, we argue that designing their point cloud counterpart is reasonable to try bridging the gap between the image and point cloud representation learning fields. However, although the previous methods generalise the convolution operation used in \acp{CNN}, they are far from closely resembling the \acp{CNN}' architecture. Specifically, \acp{CNN} adopt a sequence of pooling layers to summarise the presence of features in local patches, but the preceding approaches do not leverage such a sequence or fail to perform graph pooling around the graph's local structures. The root of this failure stems from the use of \ac{KNN}, which does not produce a point cloud’s graph with disjoint local regions. Consequently, locally down sampling a point cloud is unattainable. In this paper, we address these issues by proposing a novel unit, dubbed Shrinking unit, that converts point clouds into directed graphs by means of K-Means++, allowing us to perform graph convolution and pooling operations on disjoint local structures. As in \cite{Han.2021}, besides capturing local dependencies, the Shrinking unit also learns self correlations. Global correlations are instead explored by stacking multiple Shrinking units horizontally as in \cite{Wang.2019}. A horizontal stack of Shrinking units also allows for reducing the size of the point cloud inputs by sequentially encoding their local structures. In addition, we deploy vertical stacks to compute multiple features at a given depth of the network in the same spirit of \acp{CNN}. Despite the overwhelming success of \acp{CNN} being mainly attributed to such a generation of multiple feature maps, the previous methods fail to embody the same principle in the point cloud domain.
\subsection{Attention-based methods}
The attention mechanism is one of the most significant recent breakthroughs in deep learning research. Although it was initially developed for natural language processing tasks
\cite{Bahdanau.2014, Vaswani.2017}, its outstanding achievements have led a large number of researchers from other fields, especially computer vision \cite{Song.2022, Fan.2021}, to pay attention to it. As evinced by the gratifying results in point cloud representation learning \cite{Wan.2021, Chen.2021, Lin.2021, Wang.2019b, Feng.2020, jing2022agnet}, the attention mechanism seems to be particularly suitable also in the point cloud field. In this paper, we instead propose a framework for the design of \ac{CNN}-like 3D point cloud feature extractors. Indeed, as previously mentioned, while multiple attention-based methods have been proposed in the literature, no \ac{CNN}-like feature detector methods have been proposed for point clouds yet.

\begin{figure*}[!t]
\centering
\includegraphics[width=6.5in]{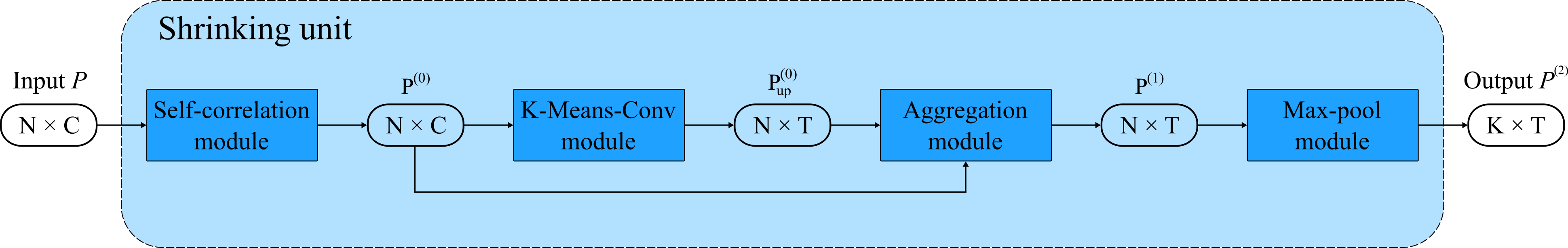}
\caption{The proposed Shrinking unit architecture. A point cloud input $P$ with $N$ $C$-dimensional points is transformed into a coarse feature point cloud $P^{(2)}$ containing $K$ $T$-dimensional points, where $T > C$ . This transformation is achieved by first piping $P$ through the Self-correlation and K-Means-Conv modules, which explore the self-correlation of each point and the local correlation between neighbouring points, respectively. Afterwards, the Aggregation module aggregates the feature resulting from the self and local correlation learning in a data-adaptive fashion. Finally, the output of the Aggregation module is fed into the Max-pool module, which produces  the coarse feature point cloud $P^{(2)}$ by applying the max-pool operation to encode every local region in $P_{out}^{(1)}$.}
\label{fig:shrinking_unit}
\end{figure*}

\section{Shrinking unit}\label{sec:shrinking_unit}
\subsection{Overview}
In this paper, we propose the Shrinking unit, a graph convolution-based unit designed to detect features in a point cloud. To accomplish this task, besides exploring self and local correlations between points, which are essential for point cloud shape understanding, this unit employs a locality-based pooling operation. That is, after self and local correlation learning, a point cloud input is shrank to a smaller set of points, each of which encodes local spatial geometric information of a point cloud input's region. These output points have a higher dimensionality than the points in the point cloud input as each of them captures geometric features of a whole local region. Therefore, the feature extraction process of the Shrinking unit can be understood as a mapping from a point cloud input to a point cloud containing fewer points of higher dimensionality. We denote the output of the Shrinking unit with the term \emph{coarse feature point cloud}. As a matter of example, let us suppose that a 3D point cloud of a table was provided as input to the Shrinking unit. A table consists of four legs and a top. The Shrinking unit would encode each of these parts with a single point of higher dimensionality ($>$3), shrinking the original point cloud to five higher-dimensional points. In fact, unless the point cloud inputs are pre-segmented, it is impractical to associate each point with a local region. For this reason, we approximate the task of local region discovery with K-Means++. The Shrinking unit is equipped with such a pooling operator to emulate the functionality of the \acp{CNN}' pooling layers.

More details about the Shrinking unit's internal working can be found in \autoref{sec:shrinking_unit_architecture}. In \autoref{sec:horiz_vert_stacks}, we discuss how emulating the \acp{CNN}' feature detection procedure to map point cloud data to feature descriptors can be accomplished by stacking multiple Shrinking units horizontally and vertically. A horizontal stack of Shrinking units also allows the exploration of global dependencies between points. Further considerations are provided in \autoref{sec:further_considerations}.

\subsection{Architecture}\label{sec:shrinking_unit_architecture}
\begin{figure}[!b]
\centering
\includegraphics[width=2.0in]{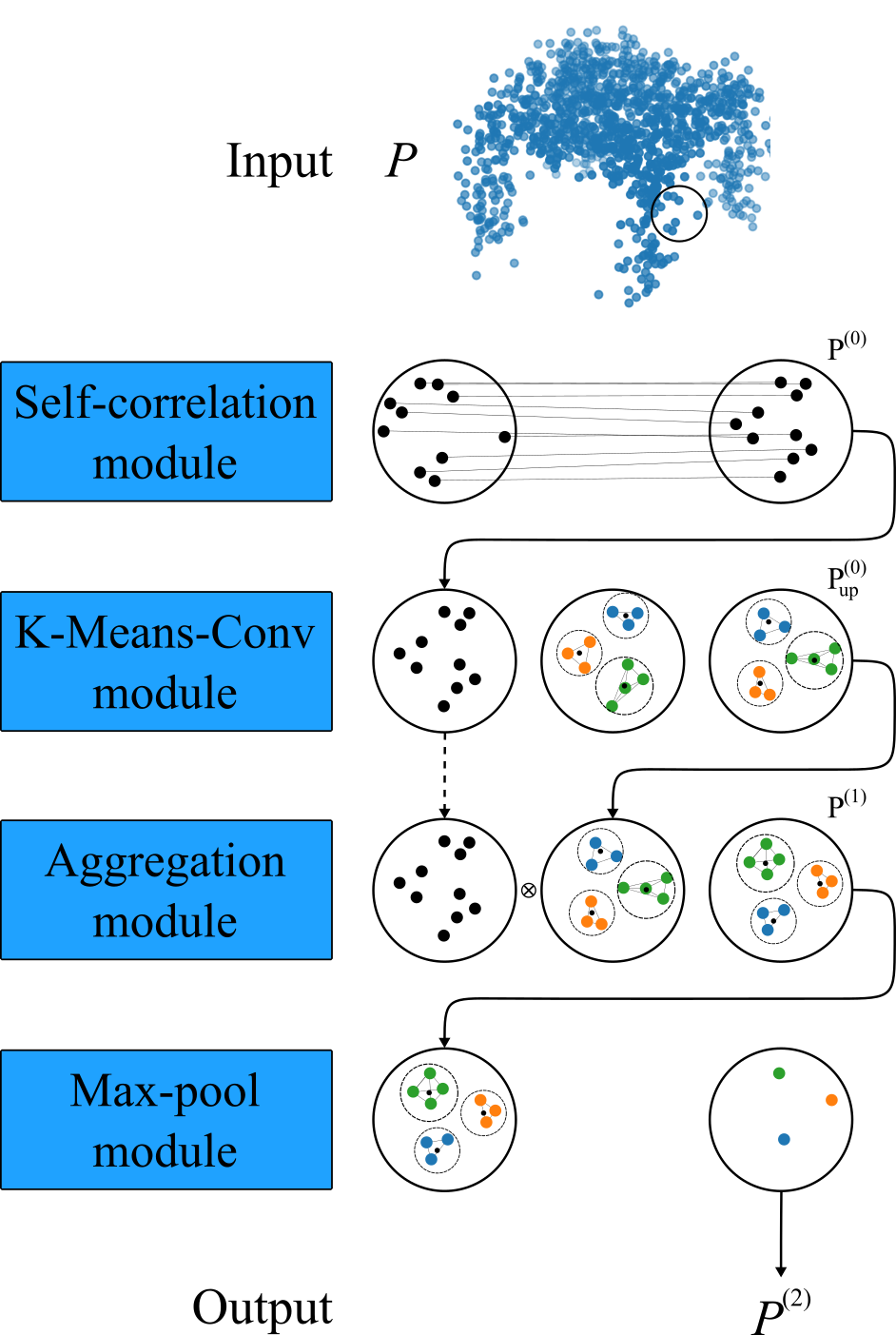}
\caption{Pictorial overview of the Shrinking unit feature detection process. For convenience, this figure provides a concrete example of how a small area of a point cloud instance gets transformed when piped through the Shrinking unit modules. First, as shown by the connections, the Self-correlation module maps each point in the small area into a new point. This mapping changes the points' arrangement and enables learning each point's self-correlation. Successively, this new arrangement of points is fed into the K-Means-Conv module. As evinced by the second and third circle next to the K-Means-Conv module block, the latter first clusters the points into three regions and successively convolves each point with respect to its cluster. This operation captures local spatial geometric information and results in a new arrangement of the clusters. Note that the output of the K-Means-Conv module is a set of higher-dimensional points, but we limit ourselves to 2D points in this figure to not overcomplicate its representation. The Aggregation module subsequently receives both the outputs of the Self-correlation module and the K-Means-Conv module and aggregates these different correlation learning features in a data-adaptive fashion. Notice that this aggregation results in a further new arrangement of the clusters. Ultimately, the Max-pool module applies the max-pooling operation to encode every cluster with a single point.}
\label{fig:shrinking_unit_illustrated}
\end{figure}

\autoref{fig:shrinking_unit} provides a schematic representation of the Shrinking unit's architecture. In contrast, \autoref{fig:shrinking_unit_illustrated} provides a pictorial overview of the operations performed by the Shrinking unit. Let $P \in R^{N \times C}$ be a point cloud containing $N$ $C$-dimensional points. The Self-correlation module receives $P$ as input,  learns the self-correlation of each point and updates them accordingly, generating a feature point cloud $P^{(0)} \in R^{N \times C}$. With the term \emph{feature point cloud}, we denote a set whose points encode some features about a given point cloud. Then, the K-Means-Conv module applies a graph convolution operation on each node to capture local dependencies in $P^{(0)}$. K-Means++ is used to compute the local receptive field of each node. Specifically, K-Means++ clusters the nodes into $K$ regions, and each node is convolved with respect to its cluster. The K-Means-Conv module outputs a feature point cloud $P^{(0)}_{up} \in R^{N \times T}$, where $T > C$. The dimensionality of the points is increased after the convolution operation because each node needs to encode the spatial geometric information of its neighbourhood as well as its own features. 

Afterwards, the Aggregation module aggregates the feature in $P^{(0)}$ and $P^{(0)}_{up}$ in a data-adaptive fashion, producing a feature point cloud $P^{(1)} \in R^{N \times T}$. Ultimately, the Max-pool module takes $P^{(1)}$ as input and generates a coarse feature point cloud $P^{(2)} \in R^{K \times T}$, a point cloud containing $K$ $T$-dimensional points. Notice that $K$ is the hyperparameter used in the K-Means-Conv module for the number of clusters. Essentially, the Max-pool module applies the max-pool operation to encode every cluster in $P^{(1)}$ with a unique point. Each cluster in $P^{(1)}$ is a descendant of a cluster previously generated in the K-Means-Conv module. In what follows, the internal working of each of the Shrinking unit's modules is more extensively analysed. Note that hereafter we adopt the notation $R^M \rightarrow R^N$ \ac{MLP} to denote an \ac{MLP} that maps an m$^{th}$-dimensional vector of real value numbers to an n$^{th}$-dimensional one.
\subsubsection{Self-correlation module}
The Self-correlation module adopts the technique proposed in \cite{Han.2021} to explore the correlation between the different channels of a node. For a given point $p_i \in P$, a channel weight $w_i \in R^C$ is computed as
\begin{equation}
    w_i = f(p_i)
\end{equation}
where $f$ is a $R^C \rightarrow R^C$ \ac{MLP}.
In order to make the scalar weights for each channel comparable, the weight vector is then normalised using \emph{softmax}
\begin{equation}
    \overline{w_{i,c}}= \frac{e^{w_{i,c}}}{\sum_{l=1}^{C}e^{w_{i,l}}}
\end{equation}
where $w_{i,c}$ is the $c^{th}$ scalar value of $w_i$ and $\overline{w_{i,c}}$ is the normalised $w_{i,c}$.
\autoref{eq:self_node_update} is then used to update the channels of a node with residual connections
\begin{equation}\label{eq:self_node_update}
    p_i^{(0)}=p_i \oplus \lambda \left(\overline{w_i} \odot p_i \right)
\end{equation}
where  $p_i^{(0)}$ is the $i^{th}$ node in $P^{(0)}$, $p_i$ is the $i^{th}$ node in $P$, $\lambda$ is a parameter to be learnt and $\oplus$, $\odot$  denote channel-wise summation and multiplication, respectively.
This update transforms an input point cloud $P \in R^{N \times C}$ into a feature point cloud $P^{(0)} \in R^{N \times C}$. Notice that this module does not use shared weights among all the nodes and thus effectively learns different correlations for each node.

\subsubsection{K-Means-Conv module}
The K-Means-Conv module explores local geometric structures in $P^{(0)}$ by applying a modified version of the graph convolution operation proposed in \cite{Valsesia.2018}. The local receptive field of each point is computed using K-Means++. Specifically, with the help of K-Means++, a directed graph $\mathcal{G}=(\mathcal{V},\mathcal{E})$ is computed from $P^{(0)}$. While $\mathcal{V}$ is the set of points in $P^{(0)}$, $\mathcal{E}$ denotes the set of directed edges, which is defined as
\begin{equation}
\begin{split}
\mathcal{E} = \{\;(&p_i^{(0)}, p_j^{(0)})\;\; |\;\; p_i^{(0)}, p_j^{(0)} \in P^{(0)} \text{ and }\\ &p_i^{(0)}, p_j^{(0)} \text{ share the same centroid}\}
\end{split}
\end{equation}
Therefore, given a choice of $K$ for the number of clusters, $\mathcal{G}$ is a collection of $K$ disjoint directed sub-graphs such that every pair of nodes in each sub-graph share the same centroid. We compute the centroids with  K-Means++ rather than the classic K-Means algorithm because its application results in less variability of the clusters between successive runs and achieves better performance overall.

Successively, the graph convolution operation in \autoref{eq:conv_operation} is applied to update each node $p_i^{(0)} \in P^{(0)}$
\begin{equation}
\begin{split}\label{eq:conv_operation}
p_{i\;\; up}^{(0)} &= \sigma \left(\;\frac{\beta}{|\mathcal{N}_i|\mathcal{M}\left(\beta \right)} + \mathcal{W}\left(p_i^{(0)}\right)p_i^{(0)} + b \; \right) \text{, with}\\
\beta &= \sum_{j \in \mathcal{N}_i}\mathcal{F}\left(p_j^{(0)} - p_i^{(0)}\right)p_j^{(0)}
\end{split}
\end{equation}
where $p_{i\;\; up}^{(0)}$ is the $i^{th}$ node in $P_{up}^{(0)}$, $p_i^{(0)}$ is the $i^{th}$ node in $P^{(0)}$, $\mathcal{N}_i$ is the neighborhood of node $p_i^{(0)}$, $\mathcal{F}$ is a $R^C \rightarrow R^T$ \ac{MLP}, $\mathcal{W}$ is a $R^C \rightarrow R^T$\ac{MLP}, $\mathcal{M}$ is a $R^T \rightarrow R$ \ac{MLP}, $b$ is a $R^{T}$ vector and $\sigma$ is the sigmoid function. This update transforms an input point cloud $P^{(0)} \in R^{N \times C}$ into a feature point cloud $P^{(0)}_{up} \in R^{N \times T}$, where $T > C$.
Note that even if two pairs of nodes are in different regions, the value   $\mathcal{F}(\cdot, \cdot)$  does not vary if their difference is identical. This reduces the number of degrees of freedom of the model, producing a form of weight sharing similar to \acp{CNN}.
\subsubsection{Aggregation module}
In order to aggregate the points in $P^{(0)}$ and $P^{(0)}_{up}$, we adopt a slightly different version of the \ac{AFA} module \cite{Han.2021}. Unlike skip connections, which linearly aggregate different features and thus are not data-adaptive,  we use a gate mechanism to make the aggregation data-aware, better representing the characteristics of the point cloud. Given $P^{(0)}$ and $P_{up}^{(0)}$, their characteristics are first described by computing their average node $s_1$ and $s_2$, respectively
\begin{align}
s_1 &= \frac{1}{N} \sum_{i=1}^N p_i^{(0)}\\
s_2 &= \frac{1}{N} \sum_{i=1}^N p_{i\;\;up}^{(0)}
\end{align}
where $p_i^{(0)}$ is the $i^{th}$ node in $P^{(0)}$ and $p_{i\;\;up}^{(0)}$ is the $i^{th}$ node in $P_{up}^{(0)}$. Then, to further explore the characteristics of the points, we map the average nodes $s_1$ and $s_2$ into new descriptors $z_1 \in R^{C}$ and $z_2 \in R^{T}$
\begin{align}
z_1 &= f_1(s_1)\\
z_2 &= f_2(s_2)
\end{align}
where $f_1$ and $f_2$ are an $R^C \rightarrow R^C$ \ac{MLP} and an $R^T \rightarrow R^T$ \ac{MLP}, respectively. To combine features in a data-aware and non-linear manner, a gate mechanism is constructed and deployed on the feature descriptors $z_1$ and $z_2$ by applying the \emph{softmax} function
\begin{align}
m_{1,c} &= \frac{e^{z_{1,c}}}{e^{z_{1,c}} + e^{z_{2,c}}}\\
m_{2,c} &= \frac{e^{z_{2,c}}}{e^{z_{1,c}} + e^{z_{2,c}}}
\end{align}
where $m_{1,c}$ and $m_{2,c}$ are the masked $z_1$ and $z_2$ on the $c^{th}$ channel and $m_{1,c} + m_{2,c}=1$. Ultimately, with the help of $m_1$ and $m_2$, $P^{(0)}$ and $P_{up}^{(0)}$ are aggregated forming $P^{(1)}$:
\begin{align}
p_i^{(1)} = m_1 \cdot p_i^{(0)} + m_2 \cdot p_{i\;\;up}^{(0)}
\end{align}
where  $p_i^{(1)}$ is the $i^{th}$ node in $P^{(1)}$, $p_i^{(0)}$ is the $i^{th}$ node in $P^{(0)}$ and $p_{i\;\;up}^{(0)}$ is the $i^{th}$ node in $P_{up}^{(0)}$. Note that $P_{up}^{(0)}$ is a feature point cloud of higher dimensionality than $P^{(0)}$, thus prior to the aggregation operation, the points in $P^{(0)}$ are augmented with zeros. This module returns $P^{(1)} \in R^{N \times T}$.
\subsubsection{Max-pool module}
The Max-pool module utilises the \emph{max-pooling} function to encode every region in $P^{(1)}$ with a single point. This is accomplished by first computing a directed graph $\mathcal{G}_2 = (\mathcal{V}_2, \mathcal{E}_2)$ from $P^{(1)}$.  $\mathcal{V}_2$ is the set of nodes in $P^{(1)}$ and $\mathcal{E}_2$ is a set of edges defined as
\begin{equation}
\begin{split}
\mathcal{E}_2 = \{\;&(p_i^{(1)}, p_j^{(1)})\;\; |\;\; p_i^{(1)}, p_j^{(1)} \in P^{(1)} \text{ and }\\ &(\text{anc}(p_i^{(1)}), \text{anc}(p_j^{(1)})) \in \mathcal{E}\;\}
\end{split}
\end{equation}
where anc$(p_i^{(1)})$ and anc$(p_j^{(1)})$ denote the ancestor node in $P^{(0)}$ of point $p_i^{(1)}$ and $p_j^{(1)}$, respectively, and $\mathcal{E}$ is the set of edges in $\mathcal{G}$, the graph constructed in the K-Means-Conv module. Therefore, $\mathcal{G}_2$ is a collection of $K$ disjoint directed sub-graphs such that for every pair of points in each sub-graph, their ancestors in $P^{(0)}$ are adjacent in $\mathcal{G}$.

The \emph{max-pooling} operation is then used to encode every sub-graph $\mathcal{G}_2^{(i)}$ in $\mathcal{G}_2$ with a single node $p_i^{(2)}$
\begin{equation}
p_i^{(2)} = max \left( \mathcal{G}_2^{(i)}\right)
\end{equation}
where $max$  is the max-pooling operation computed along each dimension of the points in $\mathcal{G}_2^{(i)}$. The Max-pool module outputs the coarse feature point cloud $P^{(2)} \in R^{K \times T}$.

\subsection{Horizontal and vertical stacks}\label{sec:horiz_vert_stacks}
The Shrinking unit is capable of exploring self and local correlations between points in a point cloud, but it cannot learn global correlations. However, as pointed out in \cite{Han.2021}, the latter kind of correlation learning is also crucial for shape understanding. Indeed, we humans unconsciously analyse the dependencies between the different parts of an object, which may be close or far apart, to understand its semantics. Furthermore, the overwhelming success of \acp{CNN} in image feature extraction suggests the value of adapting their insight to the point cloud domain. Yet, a single Shrinking unit is insufficient to increasingly reduce the size of point clouds by sequentially encoding their local structures like in \acp{CNN} as well as detect multiple features at each layer of the network. We address these issues by stacking multiple Shrinking units horizontally and vertically.

\begin{figure}[!b]
    \centering
    \begin{subfigure}{0.489\textwidth}
        \centering
        \includegraphics[width=3.4in]{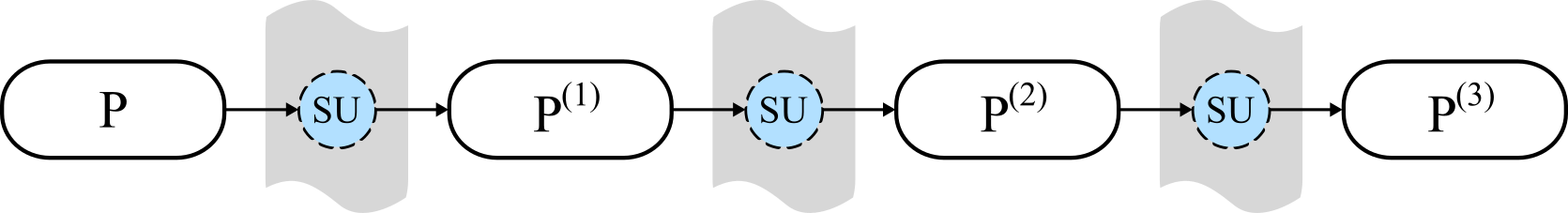}
        \subcaption{Example of three horizontally stacked Shrinking units. The blue circles denote the Shrinking units. $P$ is the point cloud input while $P^{(1)}$, $P^{(2)}$ and $P^{(3)}$ are the coarse feature point cloud outputs of the first, second and third Shrinking unit, respectively. Due to the locally-based pooling operation of the Shrinking unit, $P^{(1)}$, $P^{(2)}$ and $P^{(3)}$ increasingly contain fewer points of higher dimensionality. Additionally, while $P^{(1)}$ only encodes local structures, $P^{(2)}$ and $P^{(3)}$ progressively capture higher-level features and hence encode global correlations as well.}
        \label{fig:horizontal_stack}
    \end{subfigure}

\medskip
    \begin{subfigure}{0.489\textwidth}
        \centering
        \includegraphics[height=1.1in]{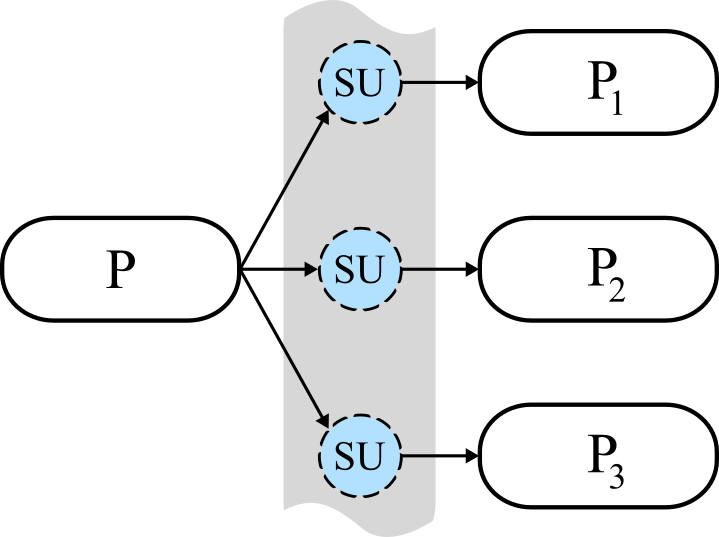}
        \subcaption{Example of three vertically stacked Shrinking units. The blue circles denote the Shrinking units. $P$ is the point cloud input while $P_1$, $P_2$ and $P_3$ are the coarse feature point cloud outputs of the first, second and third Shrinking unit, respectively. $P_1$, $P_2$ and $P_3$ encode different types of features for the same input $P$.}
        \label{fig:vertical_stack}
    \end{subfigure}
    
\medskip
    \begin{subfigure}{0.489\textwidth}
        \centering
        \includegraphics[height=1.9in]{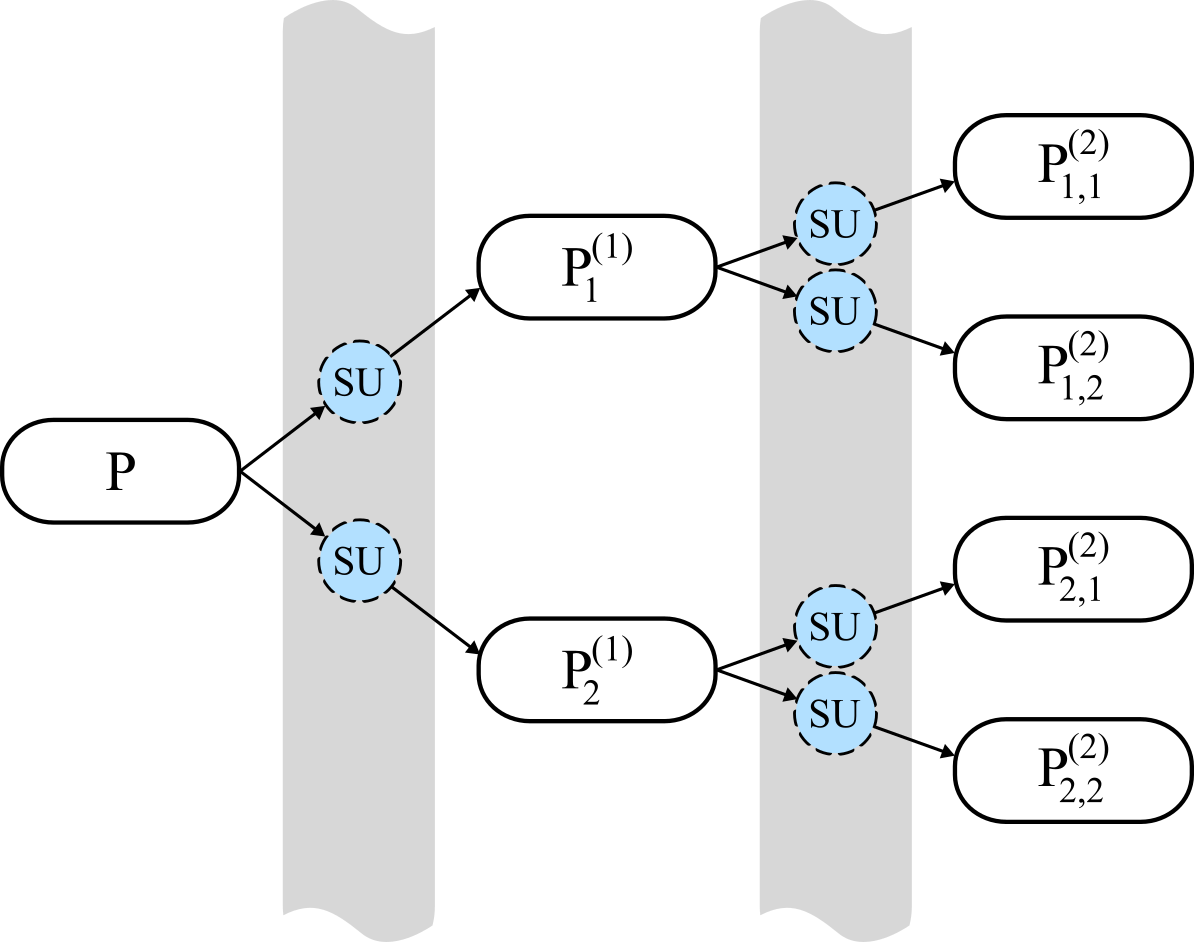}
        \subcaption{Example of horizontally and vertically stacked Shrinking units. The blue circles denote the Shrinking units. $P$, the point cloud input, is fed into two Shrinking units. The latter detect different features and output coarse feature point clouds $P_1^{(1)}$ and $P_2^{(1)}$. In turn, $P_1^{(1)}$ and $P_2^{(1)}$ are given as input to two different Shrinking units each. $P_{1,1}^{(2)}$ and $P_{1,2}^{(2)}$ capture different features from $P_1^{(1)}$, while $P_{2,1}^{(2)}$ and $P_{2,2}^{(2)}$ have different feature encodings for $P_2^{(1)}$.}
        \label{fig:shrinkingnet_shallow}
    \end{subfigure}
    \caption{Examples of horizontally and vertically stacked Shrinking units.}
\end{figure}

\paragraph{Horizontal stack}
The Shrinking unit receives a point cloud as input and outputs a coarse feature point cloud, a point cloud containing fewer points of higher dimensionality. This design enables stacking multiple Shrinking units horizontally, as shown in \autoref{fig:horizontal_stack}, such that the output of a given unit becomes the input of the next one. This enables capturing higher-level representations incrementally and thus exploring global correlations like in \cite{Wang.2019}. Also, thanks to the Max-pool module, which performs graph pooling around the point cloud local structures, a horizontal sequence of Shrinking units incrementally reduces the size of point clouds.

\paragraph{Vertical stack}
If a point cloud input is fed to multiple Shrinking units, we can detect multiple features from it. As an example, one Shrinking unit might be sensitive to spherical shapes while another might detect cubic structures from the same input. This form of multiple-feature detection greatly enhances the shape understanding process, and it has been the primary source of success of \acp{CNN} in image representation learning tasks. We extract multiple features from the same point cloud input by stacking multiple Shrinking units vertically, as shown in \autoref{fig:vertical_stack}.

Given the abovementioned properties, a combination of horizontal and vertical stacks allows us to obtain a point cloud feature extractor network that fully resembles the \acp{CNN} feature detection process as well as leverages self, local and global correlations between points. Indeed, as shown in \autoref{fig:shrinkingnet_shallow}, such a combination creates a tree structure where the root node is the point cloud input, and each branch of this tree captures a different sequence of features. Hence, multiple features are captured at each depth of the network, and the size of the point cloud inputs is incrementally reduced as well like in \acp{CNN}.

\subsection{Further considerations}\label{sec:further_considerations}
As repeatedly mentioned before, in order to build \ac{CNN}-like point cloud feature extractors, the point cloud inputs need to be increasingly reduced in size. Therefore, when stacking multiple Shrinking units vertically and horizontally, it is necessary to incrementally reduce, layer after layer, the parameter $K$ of the K-Means-Conv modules. Furthermore, the task of point cloud feature extraction typically revolves around mapping a point cloud input into a global feature descriptor vector which can then be fed into an \ac{MLP} for label assignment. As shown by \autoref{fig:shrinkingnet_shallow}, the output of a network containing vertical and horizontal stacks of Shrinking units, however, is a collection of coarse feature point clouds. In order to overcome this problem, the suggested approach is to make the last layer output coarse feature point clouds containing only a single point in a very high dimension. These points summarise different feature detection branches of the point cloud input. A symmetric aggregation function, such as max pooling, can then be applied to aggregate these points into a global feature descriptor.

\section{Experiments}\label{sec:experiments}
In this section, we demonstrate the effectiveness of our innovative idea in point cloud feature detection tasks. To this end, we build a point cloud classification network named ShrinkingNet. The latter uses multiple horizontally and vertically stacked Shrinking units for the feature detection process and then feeds the generated point clouds' global feature descriptors into an \ac{MLP} for label assignment. We evaluate the performance of ShrinkingNet on the ModelNet-10 benchmark dataset \cite{Wu.2014}. Given its relatively small size, the use of ModelNet-10 allows us to conduct experiments at a faster pace. This is of paramount importance because, as mentioned in \autoref{sec:shrinkingnet_implementation}, the training procedure of ShrinkingNet incurs a high-time complexity due to some implementation inefficiencies. As shown in \autoref{tab:method_results}, ShrinkingNet's classification accuracy (90.64\%) is comparable to the performance of the other state-of-the-art methods. Furthermore, the accuracy is very likely to increase even further if additional Shrinking units and stacked layers are added. This can be evinced by \autoref{tab:multiple_architectures}, which shows that an increase in the number of Shrinking units and stacked layers usually leads to an increase in classification accuracy. As a consequence, the \ac{CNN}-like feature detection process proposed in this paper seems to be effective. In \autoref{sec:dataset},  we give further details on dataset preprocessing. Further information on the ShrinkingNet's architecture and implementation is provided in \autoref{sec:shrinkingnet_architecture} and \autoref{sec:shrinkingnet_implementation}, respectively. \autoref{sec:training_procedure} provides details on the training procedure. Lastly, we conduct further observations on our experiments in \autoref{sec:parameters}. 

\begin{table}[]
    \centering
    \begin{tabular}{lccc}
    \hline
    Method                                           & Accuracy(\%) \\ \hline\hline
    ECC \cite{Simonovsky.2017}                       & 90.0 \\
    Achlioptas et al. \cite{Achlioptas.2018}         & 95.4 \\
    FoldingNet \cite{Yang.2018}                      & 94.4 \\ 
    KCNet \cite{Shen.2018}                           & 94.4 \\ 
    G3DNet-18 SVM, FT, Vote \cite{Dominguez.2018}    & 93.2 \\ 
    Point2Sequence  \cite{Liu.2018}                  & \textbf{95.1} \\ 
    3DCapsule \cite{Cheraghian.2019}                 & 94.7 \\ \hline
    ShrinkingNet (Ours)                              & 90.6 \\ \hline
    \end{tabular}
\caption{Classification results on the ModelNet-10 benchmark dataset. ShrinkingNet's performance is comparable to the other proposed methods.}
\label{tab:method_results}

\end{table}
\begin{table}[]
    \centering
    \resizebox{\columnwidth}{!} {%
    \begin{tabular}{lcccc}
    \hline
    Class                    & \# Shapes & Precision (\%) & Recall (\%) & F1-Score (\%) \\ \hline\hline
    bathtub                  & 50       & \textbf{100}     & 89             & 94     \\
    bed                      & 100      & 92               & 95             & 93     \\
    chair                    & 100      & \textbf{100}     & 99             & 99     \\
    desk                     & 86       & 82               & 64             & 72     \\
    dresser                  & 86       & 75               & 81             & 78     \\
    monitor                  & 100      & \textbf{100}     & 96             & 98     \\
    night stand              & 86       & 88               & 79             & 83     \\
    sofa                     & 100      & 98               & 97             & 97     \\
    table                    & 100      & 77               & \textbf{100}   & 87     \\
    toilet                   & 100      & \textbf{100}     & 93             & 97     \\ \hline
    %macro average           &          & 91               & 89             & 90     \\
    \end{tabular}
    }
\caption{Detailed ShrinkingNet's classification results on the ModelNet-10 benchmark dataset.}
\label{tab:detailed_results}
\end{table}

\subsection{Dataset preprocessing}\label{sec:dataset}
We evaluate ShrinkingNet's performance on the ModelNet-10 benchmark dataset. This dataset contains 4,899 untextured CAD models (3,991 for training and 908 for testing) divided into 10 categories. 
In order to convert these 3D meshes into 3D point clouds, we sample a fixed number of points $S=1200$ on the mesh faces according to their area, obtaining a set of 3D Euclidean coordinates. Ultimately, due to a significant difference in the intra-class and inter-class model dimensions, we apply data normalisation to scale each point dimension in the interval $[-1, 1]$.

\begin{figure}[!t]
\centering
\includegraphics[width=3.45in]{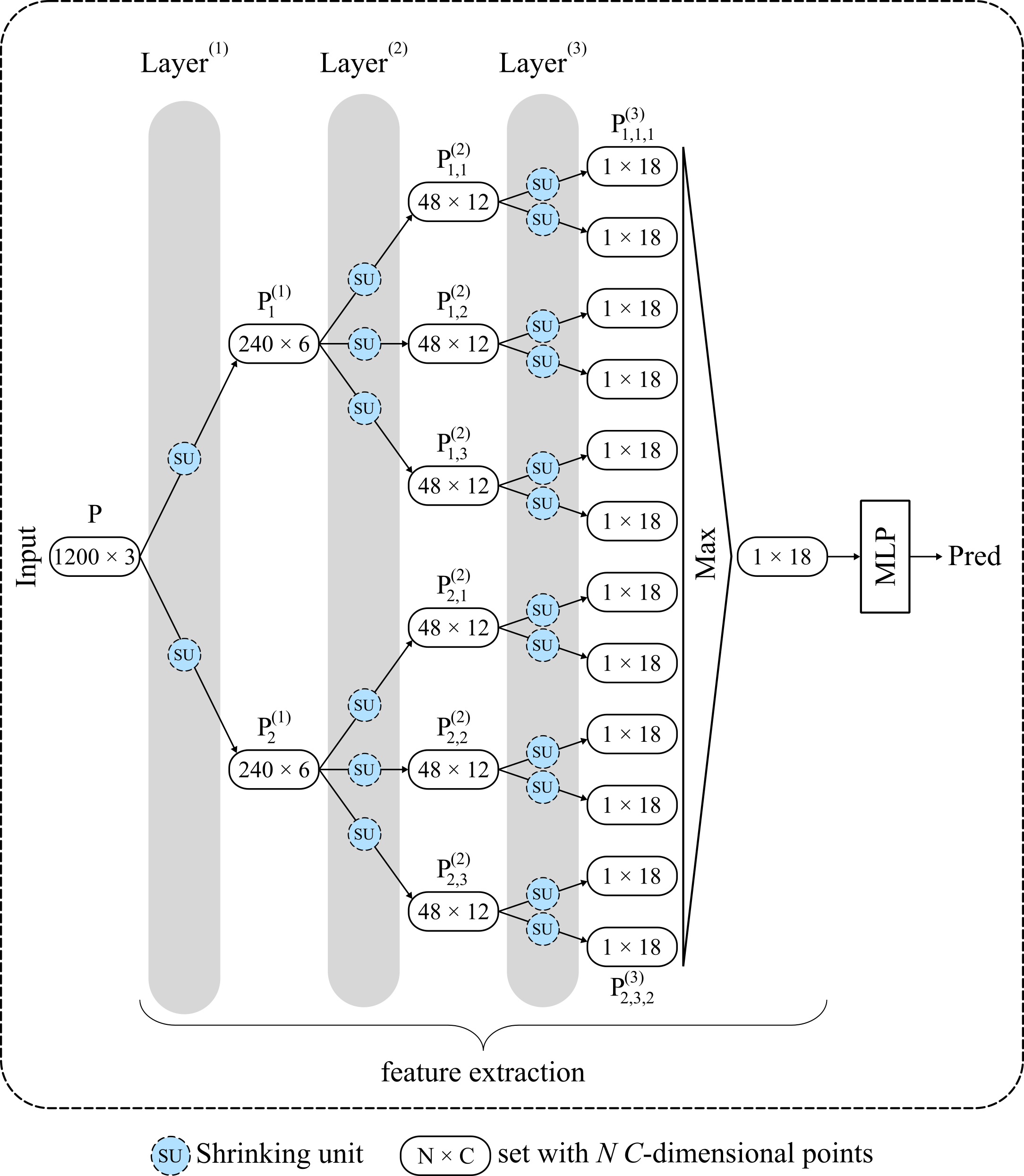}
\caption{Graphical representation of the ShrinkingNet's architecture. A point cloud input $P \in R^{1200 \times 3}$ is transformed by two Shrinking units, resulting in two different but equal-sized coarse feature point clouds $P_1^{(1)} \in R^{240 \times 6}$ and $P_2^{(1)} \in R^{240 \times 6}$. $P_1^{(1)}$ and $P_2^{(1)}$ are then processed by three Shrinking units each. The outputs of this processing, six coarse feature point clouds, are in turn processed by two Shrinking units each. Ultimately, the max-pooling operation is applied to aggregate the outputs of the third stack into a global feature descriptor. The latter is fed into an \ac{MLP} for label assignment.}
\label{fig:shrinking_net}
\end{figure}

\begin{table*}[!t]
    \centering
    \resizebox{\textwidth}{!} {%
    \begin{tabular}{ c|c|c|c } 
        \hline
          Layers & Self-correlation & K-Means-Conv & Aggregation\\
        \hline
        \hline
        \multirow{4}{*}{$1^{st}$ Layer} &  & $K=240$ & \\ 
                              & $f=$ MLP(\textcolor{blue}{3}, 13, 18, 23, 18, 13, \textcolor{cyan}{3}) & $\mathcal{F}=$ MLP(\textcolor{blue}{3}, 13, 18, 23, 18, 13, \textcolor{cyan}{6}) & $f_1=$ MLP(\textcolor{blue}{6}, 16, 21, 26, 21, 16, \textcolor{cyan}{6})\\
                              &                                    & $\mathcal{M}=$ MLP(\textcolor{blue}{6}, 16, 21, 26, 21, 16, \textcolor{cyan}{1}) & $f_2=$ MLP(\textcolor{blue}{6}, 16, 21, 26, 21, 16, \textcolor{cyan}{6}) \\ 
                              &                                    & $\mathcal{W}=$ MLP(\textcolor{blue}{3}, 13, 18, 23, 18, 13, \textcolor{cyan}{6}) & \\

        \hline
        \hline
        \multirow{4}{*}{$2^{nd}$ Layer} &  & $K=48$ &  \\
                              & $f=$ MLP(\textcolor{blue}{6}, 16, 21, 26, 21, 16, \textcolor{cyan}{6})  & $\mathcal{F}=$ MLP(\textcolor{blue}{6}, 16, 21, 26, 21, 16, \textcolor{cyan}{12}) & $f_1=$ MLP(\textcolor{blue}{12}, 22, 27, 32, 27, 22, \textcolor{cyan}{12})\\ 
                              &                                    & $\mathcal{M}=$ MLP(\textcolor{blue}{12}, 22, 27, 32, 27, 22, \textcolor{cyan}{1}) & $f_2=$ MLP(\textcolor{blue}{12}, 22, 27, 32, 27, 22, \textcolor{cyan}{12}) \\
                              &                                    & $\mathcal{W}=$MLP(\textcolor{blue}{6}, 16, 21, 26, 21, 16, \textcolor{cyan}{12}) & \\ 
        \hline          
        \hline
        \multirow{4}{*}{$3^{rd}$ Layer} &  & $K=1$ &  \\
                              & $f=$ MLP(\textcolor{blue}{12}, 22, 27, 32, 27, 22, \textcolor{cyan}{12}) & $\mathcal{F}=$ MLP(\textcolor{blue}{12}, 22, 27, 32, 27, 22, \textcolor{cyan}{18}) & $f_1=$ MLP(\textcolor{blue}{18}, 28, 33, 38, 33, 28, \textcolor{cyan}{18})\\
                              &                                     & $\mathcal{M}=$ MLP(\textcolor{blue}{18}, 28, 33, 38, 33, 28, \textcolor{cyan}{1})  &  $f_2=$ MLP(\textcolor{blue}{18}, 28, 33, 38, 33, 28, \textcolor{cyan}{18})\\
                              &                                     & $\mathcal{W}=$ MLP(\textcolor{blue}{12}, 22, 27, 32, 27, 22, \textcolor{cyan}{18}) & \\
        \hline
        
    \end{tabular}
    }
    \caption{Parameters and \acp{MLP} used in the Shrinking units of ShrinkingNet. Each row shows the parameters and \acp{MLP} of the Shrinking units in the corresponding ShrinkingNet's layer. The notation MLP$(m_1, ..., m_n)$ denotes an \ac{MLP} with $n$ layers whose number of neurons is $m_1,...,m_n$, respectively. The \textcolor{cyan}{cyan-coloured} and \textcolor{blue}{blue-coloured} digits denote the number of neurons in the input and output layers, respectively. The activation function of each neuron is \ac{ReLU}.}
    \label{tab:extendingNet_shapenet}
\end{table*}

\subsection{ShrinkingNet's architecture}\label{sec:shrinkingnet_architecture}
\autoref{fig:shrinking_net} illustrates the ShrinkingNet's architecture. 
ShrinkingNet consists of three vertical stacks of Shrinking units. The first stack comprises two Shrinking units. The latter take as input a point cloud input $P \in R^{1200 \times 3}$ and output a coarse feature point cloud $P_1^{(1)} \in R^{240 \times 6}$ and $P_2^{(1)} \in R^{240 \times 6}$, respectively. Thus, the first stack shrinks the initial point cloud to two different sets of 240 six-dimensional points. In the second layer, $P_1^{(1)}$ and $P_2^{(1)}$ are then processed by three different Shrinking units each. This results in six different coarse feature point clouds containing 48 twelve-dimensional points. The latter are successively fed into the third stack. The third stack contains twelve Shrinking units, two units for each input, and thus outputs twelve coarse feature point clouds. These coarse feature point clouds contain a single eighteen-dimensional point. Upon aggregating the outputs of the third stack into a single vector by applying the max-pooling operation along each of their dimension, the generated global feature descriptor is fed into an \ac{MLP} for label assignment. This \ac{MLP} comprises six layers of 18, 36, 46, 56, 46 and 10 neurons, respectively. The activation function of each neuron is \ac{ReLU}. \autoref{tab:extendingNet_shapenet} provides further details on the parameters and \acp{MLP} used inside the Shrinking units of ShrinkingNet.

\subsection{ShrinkingNet's implementation}\label{sec:shrinkingnet_implementation}
ShrinkingNet is implemented using the PyTorch \cite{paszke2019pytorch} and PyTorch Geometric \cite{fey2019fast} libraries. The latter allows writing geometric deep learning structures and operations efficiently. Therefore, it is particularly suitable for the development of ShrinkingNet. In addition, every operation based on PyTorch Geometric code is automatically provided with multi-GPU support. As a consequence, the training processes can be sped up in multi-GPU environments.

Implementing the ShrinkingNet's architecture mainly revolves around coding the Shrinking unit and stacking it vertically and horizontally. The Shrinking unit subclasses the torch.nn.Module class, making it easy to plug the unit into existing architectures to improve their performance. Its implementation cannot be improved further. In contrast, the currently adopted approach to stack multiple Shrinking units vertically is by no means efficient. Indeed, it is based on Python multithreading programming, while the proper solution requires writing multithreaded code in native C/C++ and CUDA for the CPU and GPU environments, respectively, instead. Our code is available at \href{https://github.com/albertotamajo/Shrinking-unit}{github.com/albertotamajo/Shrinking-unit}
\subsection{Training procedure}\label{sec:training_procedure}
During the experimental phase, we trained our model on four NVIDIA DGX GPUs. For each training process, the network's weights were initialised using the Glorot initialisation technique \cite{Glorot.2010}. The negative log-likelihood loss function, a batch size of 32, and the Adam solver with a learning rate of 3e-4 and a momentum of 0.999 were used to update the network's parameters. Weight decay for L2 penalty remained unconsidered. After each epoch, a data augmentation technique, noise addition, was used to increase the diversity of the training dataset so that to reduce overfitting and achieve better generalisation performance.
We trained our model for at most 200 epochs, adopting an early stopping technique in case of overfitting or convergence.

\begin{table}[!t]
    \centering
    \begin{tabular}{cccc}
    \hline
    \# Layers & \# Shrinking units & Best Accuracy(\%) \\ \hline\hline
    2 & $[1, 1]$                                   & 50.26 \\
    3 & $[1, 1, 1]$                                & 69.78 \\
    3 & $[2, 4, 8]$                                & 76.76 \\ \hline
    3 & $[2, 6, 12]$                               & \textbf{90.64} \\
    \hline
    \end{tabular}
\caption{Best classification accuracy on ModelNet-10 among all the models we trained with the same number of layers and Shrinking units. While the first column indicates the number of layers, the second column shows the number of Shrinking units contained in each layer. As clearly evinced, the classification accuracy improves as the number of layers and Shrinking units increases. The highest classification accuracy (90.64\%) is achieved by ShrinkingNet.}
\label{tab:multiple_architectures}
\end{table}

\subsection{Further observations}\label{sec:parameters}
Besides ShrinkingNet, we constructed and trained several point cloud classification architectures equipped with a different number of Shrinking units and stacked layers for the feature detection process. In \autoref{tab:multiple_architectures}, we can observe that an increase in the number of Shrinking units and stacked layers generally improves classification accuracy on ModelNet-10. This is a crucial observation because it suggests that deeper networks comprising a larger number of Shrinking units and stacked layers, in the same spirit as current \acp{CNN}, have the potential to achieve state-of-the-art performance. 

\begin{figure}[!t]
\centering
\includegraphics[width=3.45in]{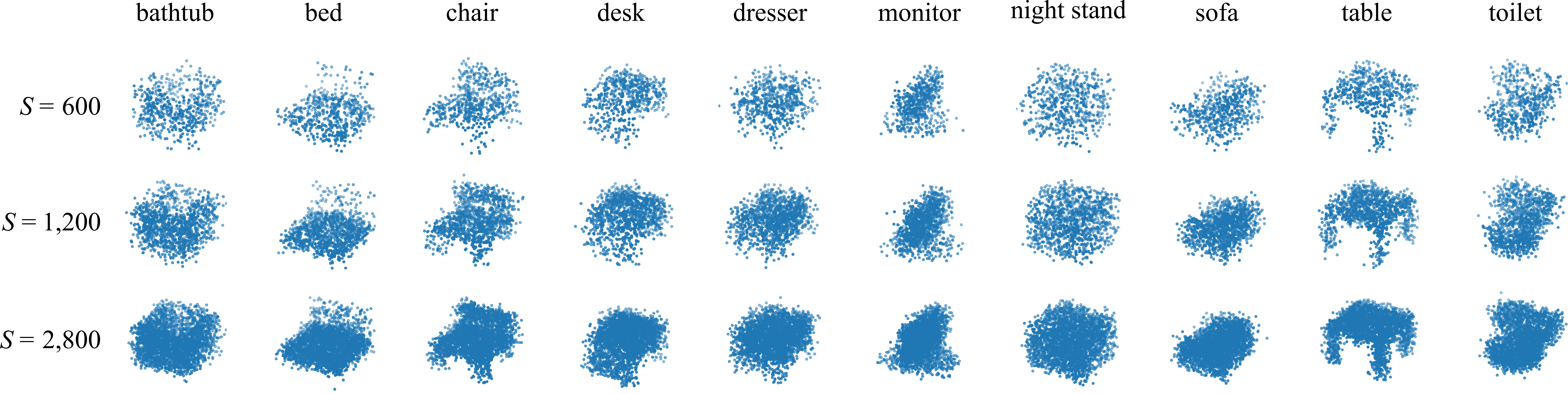}
\caption{Visualisation of randomly selected ModelNet-10 CAD models. The CAD models are converted into 3D point clouds by sampling a fixed number of points $S$ on the mesh faces according to their area.}
\label{fig:sampled_data}
\end{figure}

Hereafter, we observe how changing some parameters in the ShrinkingNet's architecture or training procedure affects its performance. First, we explore the effect of the number of sample points $S$, which affects the density of points clouds as shown in \autoref{fig:sampled_data},  on the ShrinkingNet's classification accuracy.
\autoref{tab:sampling_influence} shows that the accuracy rises between $S=600$ and $S=1200$ but ultimately it drops when $S=2800$. This implies that ShrinkingNet is sensitive to different point cloud densities, and $S = 1,200$ is the optimal sampling density to capture the features in the ModelNet-10 models. 
\begin{table}[!b]
    \centering
    \begin{tabular}{lccc}
    \hline
    $S$      & 600    & 1,200           & 2,800  \\ \hline
    Acc (\%) & 84.47  & \textbf{90.64}  & 87.89
    \end{tabular}
\caption{The effect of the number of sample points $S$ on ShrinkingNet's performance on the ModelNet-10 benchmark dataset.}
\label{tab:sampling_influence}
\end{table}

Then, we show the effect of augmenting the point cloud inputs with vector normals, increasing the dimensionality from 3 to 6.
As shown in \autoref{tab:dimensionality_influence}, this extra information does not help our network in the classification task. In fact, the classification accuracy slightly decreases to 86.94\%.

\begin{table}[]
    \centering
    \begin{tabular}{lccc}
    \hline
    Dim      & 3              & 6  \\ \hline
    Acc (\%) & \textbf{90.64} & 86.94 
    \end{tabular}
\caption{The effect of augmenting the inputs with vector normals, increasing their dimensionality from 3 to 6, on ShrinkingNet's performance on the ModelNet-10 benchmark dataset.}
\label{tab:dimensionality_influence}
\end{table}

\begin{table}[h]
    \centering
    \begin{tabular}{lccc}
    \hline
    Dims     & [6,9,12] & [6,12,18]        & [6,12,21]  \\ \hline
    Acc (\%) & 83.04    & \textbf{90.64}    & 76.87
    \end{tabular}
\caption{The effect of the layer-wise coarse feature point cloud dimensionalities on ShrinkingNet's performance on the ModelNet-10 benchmark dataset.}
\label{tab:p_value_influence}
\end{table}

\begin{table}[!t]
    \centering
    \begin{tabular}{lccc}
    \hline
    $LR$      & 1$e^{-2}$    & 1$e^{-3}$  & 3$e^{-4}$  \\ \hline
    Acc (\%)  & 84.47        & 83.26      & \textbf{90.64}
    \end{tabular}
\caption{The effect of the learning rate $LR$ on ShrinkingNet's performance.}
\label{tab:learninig_rate_influence}
\end{table}

Moreover, we also investigate the effect of the layer-wise coarse feature point cloud dimensionalities. For this purpose, we successively increase the dimensionality of the layer outputs from $[6, 9, 12]$ to $[6, 12, 21]$. \autoref{tab:p_value_influence} shows the results of this experiment and demonstrates that the $[6, 12, 18]$ setting, the one used in the ShrinkingNet architecture, is optimal. These results suggest that the dimensionality of the layer outputs needs to be increased gradually, layer after layer. Thus,  if the network needs to output coarse feature point clouds of very high dimensionality, additional layers need to be stacked. 

Finally, we study the effect of the learning rate (LR) on ShrinkingNet's performance. To this end, we set the LR to 1$e^{-2}$, 1$e^{-3}$, and 3$e^{-4}$. \autoref{tab:learninig_rate_influence} shows that the highest accuracy is reached at the smallest LR. This effect is not surprising, although it inevitably leads to a longer training.

\section{Conclusion and Outlook}\label{sec:conclusion}
In this work, we design a graph convolution-based unit, dubbed Shrinking unit, to detect features in point clouds. After leveraging self and local correlations between points, this unit adopts a locality-based pooling procedure to shrink point cloud inputs into coarse feature point clouds. This allows for reducing the number of points in the initial point clouds by summarising local spatial geometric information. Since \acp{CNN} achieve outstanding performance in image detection tasks, we argue that designing their point cloud counterparts deserves an attempt. For this reason, we also propose to stack multiple Shrinking units vertically and horizontally in a multilayer fashion for the development of the first \ac{CNN}-like 3D point cloud feature extractors. While a horizontal stack of Shrinking units allows for an incremental reduction in the point cloud size by progressively encoding local structures, vertical stacks enable the detection of multiple features at a given depth of the network. A horizontal stack also allows for the exploration of global correlations, which are also crucial for shape understanding. Our experiments suggest that our innovative idea of constructing \ac{CNN}-like 3D point cloud feature extractors made up of multiple Shrinking units is effective. Indeed, we achieve significant classification accuracy (90.64\%) on the ModelNet-10 benchmark dataset. Furthermore, this performance is very likely to increase even further if additional Shrinking units and stacked layers are added. This can be evinced by our results, which show that an increase in the number of Shrinking units and stacked layers generally leads to an increase in classification accuracy. In light of these results, we reckon that the \ac{CNN}-like feature extraction framework proposed in this paper has the potential to bridge the gap between the point cloud and image representation learning fields. Therefore, we urge the research community to conduct further research in this regard. First, we suggest designing deeper networks and testing their performance on large-scale point cloud datasets. Last but not least, we highly recommend developing alternative feature extraction units that can be stacked vertically and horizontally as the Shrinking unit so that to maximise the likelihood of aligning the point cloud feature extraction performance with the image one.

%%%%%%%%%%%% Supplementary Methods %%%%%%%%%%%%
%\footnotesize
%\section*{Methods}

%%%%%%%%%%%%% Acknowledgements %%%%%%%%%%%%%
\footnotesize
\section*{Acknowledgements}
The authors would like to thank the Kaiserslautern Regional University Computer Center for providing access to the high-performance cluster $Elwetritsch$ and its corresponding usage support.

%%%%%%%%%%%%%%   Bibliography   %%%%%%%%%%%%%%
\normalsize
\bibliography{refs}

%%%%%%%%%%%%  Supplementary Figures  %%%%%%%%%%%%
%\clearpage

%%%%%%%%%%%%%%%%   End   %%%%%%%%%%%%%%%%
%\end{multicols}  % Method B for two-column formatting (doesn't play well with line numbers), comment out if using method A
\end{document}